\documentclass{llncs}
\usepackage{graphicx}
\usepackage{algorithm}
\usepackage{algorithmic}

\usepackage{array}
\newcolumntype{L}[1]{>{\raggedright\let\newline\\\arraybackslash\hspace{0pt}}m{#1}}
\newcolumntype{C}[1]{>{\centering\let\newline\\\arraybackslash\hspace{0pt}}m{#1}}
\newcolumntype{R}[1]{>{\raggedleft\let\newline\\\arraybackslash\hspace{0pt}}m{#1}}

\usepackage{tabularx}
\begin{document}

\title{Conjunctive Query Based Constraint Solving For \\Feature Model Configuration}
\titlerunning{Conjunctive Query Based Constraint Solving For FM Configuration}
%
\author{Alexander Felfernig\orcidID{0000-0003-0108-3146}
\and
Viet-Man Le\orcidID{0000-0001-5778-975X}
\and
Sebastian Lubos\orcidID{0000-0002-5024-3786}
}
\authorrunning{A. Felfernig et al.}
%
\institute{Institute of Software Technology, Graz University of Technology, Graz, Austria
\email{\{alexander.felfernig,vietman.le,slubos\}@ist.tugraz.at}}
\maketitle              

\begin{abstract}
Feature model configuration can be supported on the basis of various types of reasoning approaches. Examples thereof are SAT solving, constraint solving, and answer set programming (ASP). Using these approaches requires technical expertise of how to define and solve the underlying configuration problem. In this paper, we show how to apply conjunctive queries typically supported by today's relational database systems to solve constraint satisfaction problems (CSP) and -- more specifically -- feature model configuration tasks. This approach allows the application of a wide-spread database technology to solve configuration tasks and also allows for new algorithmic approaches when it comes to the identification and resolution of inconsistencies.

\end{abstract}

\section{Introduction}\label{introduction}

Feature models (FM) can be used to represent commonality and variability aspects of highly-variant software and hardware systems \cite{Acher2018,Apel2009,BeSeRu2010,FelfernigVamos2021,Kang1990}. These models can be  specified on a graphical level and then translated into a corresponding formal representation that allows for feature model configuration including associated reasoning tasks such as conflict detection (e.g., induced by user requirements \cite{Junker2004}) and conflict resolution (e.g., resolving the existing conflicts in a void feature model \cite{Bakker93,Hentze2021Vamos,LeICSE2021,Reiter1987,White2010}).

There are different reasoning approaches supporting feature model configuration \cite{Felfernigetal2014,PopescuJIIS2022}. First, SAT solving \cite{GuPuFrWa1996} allows for a direct translation of individual features into a corresponding set of Boolean variables specifying feature inclusion or exclusion. SAT solving has shown to be an efficient reasoning approach supporting feature model analysis and configuration \cite{Mendon2009}. Second, constraint satisfaction problems (CSPs) can be used to represent feature model configuration tasks \cite{Benavides2005Constraints}. Compared to SAT solving, CSPs  allow more ``direct'' constraint representations expressing typical constraint types such as logical equivalence and implications. Finally, answer set programming (ASP) supports  the definition of object-centered configuration problems \cite{Myllaerniemi2014}. On the reasoning level, these problems are translated into a corresponding Boolean Satisfiability (SAT) based representation which can then be used by a SAT solver.

All of the afore mentioned knowledge representations require additional technical expertise in representing feature model configuration knowledge and identifying corresponding configurations (on the reasoning level either in the form of SAT solving, constraint solving, or answer set programming). Furthermore, the resulting knowledge bases have to be included using the corresponding application programming interfaces. An alternative to SAT solving, constraint satisfaction problems, and ASP is to represent feature model configuration problems in terms of \emph{conjunctive queries}. 

The major contributions of this paper are the following: (a) we show how conjunctive queries can be used to represent and solve configuration tasks, (b) we report initial results of a corresponding performance evaluation, and (c) we sketch how such approaches can be exploited to make related consistency management tasks (e.g., feature model diagnosis) more efficient. 

The remainder of this paper is organized as follows. In Section \ref{sec:examplefeaturemodel}, we introduce an example feature model that is used as working example throughout this paper. In Section \ref{sec:conjunctivequerybasedconfiguration}, we show how configuration tasks can be supported on the basis of conjunctive queries and introduce corresponding definitions of a configuration task and a feature model configuration. Alternative approaches to conflict  resolution are discussed in Section \ref{sec:advancedconflictresolution}.  Thereafter, we present initial results of a performance analysis that compares the efficiency of conjunctive query based configuration with constraint solving (see Section \ref{sec:performanceanalysis}). Section \ref{sec:validitythreats} includes a discussion of potential threats to validity. Finally, in Section \ref{sec:conclusionsandresearchissues} we conclude the paper with a discussion of open issues for future work.

\section{Example Feature Model}\label{sec:examplefeaturemodel}

\begin{figure}[ht]  
    \centering 
    \fbox{ 
        \includegraphics[width=0.6 \textwidth]{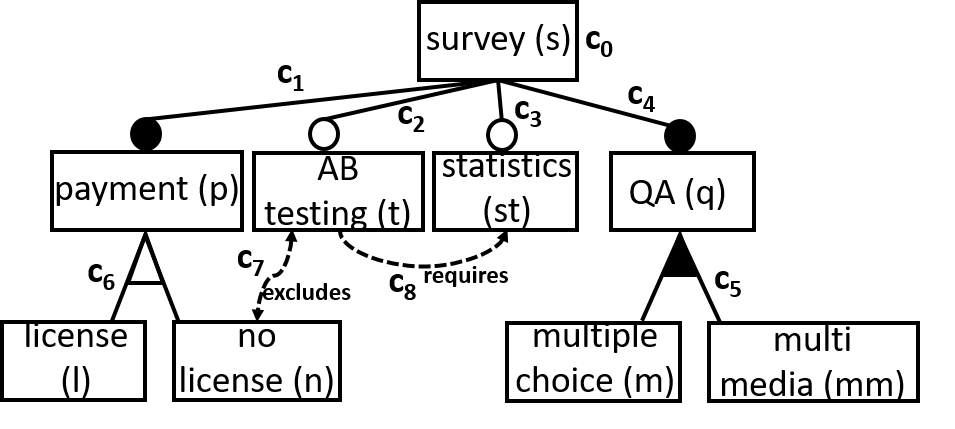}
    }
    \caption{An example feature model (\emph{survey software}).} \label{fig:FMSurveySoftware}
\end{figure}

For demonstration purposes, we introduce a simplified \emph{survey software} feature model (see Figure \ref{fig:FMSurveySoftware}). Typically, features in such models are organized in a hierarchical fashion \cite{BeSeRu2010} where relationships between features specify the overall model structure: (a) \emph{mandatory} relationships indicate that specific features have to be included in a configuration, for example, each survey software must include a corresponding payment, (b) \emph{optional} relationships indicate that specific features can be included in a configuration, for example, the AB testing feature is regarded as optional in a survey software configuration, (c) \emph{alternative} relationships indicate that exactly one of a given set of subfeatures must be included in a configuration (if the parent feature is included), for example, a payment is either a license or a no license, (d) \emph{or} relationships indicate the optional inclusion of features from a given set of subfeatures given that the parent feature is included in the configuration, for example, questions (feature QA) can additionally include (beside single-choice) multiple-choice and multimedia-style questions.

Furthermore, so-called cross-tree constraints specify additional properties which are orthogonal to the discussed relationships: (a) \emph{excludes} relationships between two features specify that the two features must not be included at the same time, for example, the \emph{no license} feature does not allow the inclusion of AB testing and vice-versa, (b) \emph{requires} relationships between two features specify the fact that if a specific feature $f_1$ is included, then another feature $f_2$ must be included, i.e., $f_1$ requires $f_2$. For example, the inclusion of AB testing  requires the inclusion of statistics.

To support reasoning operations on feature model properties, these models have to be translated into a formal representation. Examples of such representations are SAT problems \cite{Gomes2008,Mendon2009}, constraint satisfaction problems \cite{Benavidesetal2005,Rossi2006}, and answer set programs \cite{Myllaerniemi2014}. In the following, we show how a feature model configuration task can be represented and solved on the basis of  \emph{conjunctive queries}.

\section{Conjunctive Query Based Configuration}\label{sec:conjunctivequerybasedconfiguration}

Representing a feature model configuration task in terms of a conjunctive query allows the application of relational database technologies to support configuration tasks. In the following, we introduce the concepts of a feature model configuration task (see Definition \ref{def1}) and a feature model configuration (see Definition \ref{def2}) on the basis of a conjunctive query $F_{[c]}S$ where $F$ can be (a) a table representing all possible feature model configurations (i.e., an enumeration of all possible feature model configurations), (b) a set of individual tables where each table represents an individual feature from the feature model, and (c) a set of individual tables representing tuples consistent with individual constraints of the feature model. Furthermore, $[c]$ is a set of criteria that have to be fulfilled -- this includes feature model constraints  as well as requirements defined by a configurator user (customer). Finally, $S$ represents a so-called projection, i.e., those attributes (and values) that should be included in a query result.

\begin{definition}[FM Configuration Task]\label{def1}
\emph{A feature model (FM) \emph{configuration task} can be defined as conjunctive query $F_{[c]}S$ where $F$ represents the available features in tabular form (explicitly or implicitly), $[c]$ is the selection part of a conjunctive query representing criteria to be taken into account by the feature model configuration, and $S$ is a set of projection attributes (features), i.e., features whose settings should be included as a part of the shown configuration result. Furthermore, $c=cr \cup cf$ where $cr$ represents a set of customer requirements (i.e., feature inclusions and exclusions) and $cf$ represents a set of constraints derived from the feature model.}
\end{definition}

In our working example, $cf=\{c_0 .. c_8\}$ ($c_0$ is the root constraint assuming that we are not interested in empty feature models). Furthermore, if we assume that the user of the feature model configurator is not interested in paying licenses and not interested in a multimedia based question support, $cr=[s=1,p=1,n=1,mm=0]$. The translation of a feature model representation into a corresponding conjunctive query based representation is shown in Table \ref{tab:constraintsconjunctivequery}. For simplicity, in this formalization we assume a single-table representation which includes a complete enumeration of the whole configuration space.

\begin{table*}
\centering 
\caption{Translating feature model constraints into queries ($cf=\{c_0..c_8\}$).}
\scriptsize
\begin{tabular}{|c|c|} 
\hline
  constraint                 &  conjunctive query based representation       
\tabularnewline
\hline \hline
$c_0$                      &   $F_{[s=1]}\{s .. mm\}$             
\tabularnewline
\hline
$c_1$                      &   $F_{[s=1 \land p = 1 \lor s=0 \land p=0]}\{s .. mm\}$            
\tabularnewline
\hline
$c_2$                    &   $F_{[\neg t=1 \lor s=1]}\{s .. mm\}$              
\tabularnewline
\hline
$c_3$                  &   $F_{[\neg st=1 \lor s=1]}\{s .. mm\}$            
\tabularnewline
\hline
$c_4$                  &   $F_{[s=1 \land q = 1 \lor s=0 \land q=0]}\{s .. mm\}$            
\tabularnewline
\hline
$c_5$                  &   $F_{[q=1 \land (m = 1 \lor mm=1) \lor q=0 \land m = 0 \land mm=0]}\{s .. mm\}$            
\tabularnewline
\hline
$c_6$           &   $F_{[ ((\neg l=1 \lor n=0 \land p=1) \land 
 (\neg (n=0 \land p=1) \lor l=1))  
  \land ((\neg n=1 \lor l=0 \land p=1) \land   
(\neg (l=0  \land p=1) \lor n=1))   ]}\{s .. mm\}$  
\tabularnewline
\hline
$c_7$            &       $F_{[\neg t=1 \lor  \neg n=1]}\{s .. mm\}$   
\tabularnewline
\hline
$c_8$            &       $F_{[\neg t=1 \lor st=1]}\{s .. mm\}$             
\tabularnewline
\hline
\end{tabular} 
\label{tab:constraintsconjunctivequery} 
\end{table*}

Given the definition of a feature model configuration task $F_{[c]}S$, we are able to introduce the definition of a corresponding feature model configuration (see Definition \ref{def2}).

\begin{definition}[FM Configuration]\label{def2}
\emph{A feature model (FM) \emph{configuration} for a FM configuration task $F_{[c]}S$ is a tuple representing a result of executing the corresponding conjunctive query.}
\end{definition}

Following the ``single table'' approach which follows the idea of explicitly representing the whole configuration space, an example of a configuration result is the tuple \{$s=1, p=1,l=0,n=1,t=0,st=1,q=1,m=1,mm=0$\}.

\emph{(a) All Possible Feature Model Configurations}. One possibility to represent the feature configuration space is to explicitly enumerate all possible configurations\footnote{This enumeration can also be performed on the basis of a conjunctive query where [c] represents the set of feature model constraints.} and regard feature model configuration as basic query on a single table $F$ (see Table \ref{tab:tableScompleteenumeration}). Such a representation can make sense only if the configuration space is small and corresponding conjunctive queries on $F$ can be executed in an efficient fashion. In this context, a conjunctive query returning a configuration could be the following: $F_{[s=1,p=1,n=1,mm=0]}\{s,p,l,n,t,st,q,m,mm\}$ resulting, for example, in a  feature model configuration $\{s=1,p=1,l=0,n=1,t=0,st=1,q=1,m=1,mm=0\}$. On the level of the structured query language (\textsc{SQL}), a related query is: \textsc{SELECT} \emph{s,p,l,n,t,st,q,m,mm} \textsc{FROM} \emph{F} \textsc{WHERE} \emph{s=1 and p=1 and n=1 and mm=0}.\footnote{To assure that only one tuple is returned as configuration, we assume a query setting such as \emph{LIMIT=1} (this is database-specific).} 

\emph{Example Query Optimizations}. A possibility of improving the performance of queries on $F$ is to reduce the number of tuples in $F$ by assuming that the root feature in $F$ (in our case, the feature \emph{survey} (\emph{s})) is always \emph{true} ($1$).

\begin{table}[ht]
\centering  
\caption{Explicit configuration space description in one table $F$ including the attributes $s$ .. $mm$.} 
\begin{tabular}{|c|c|c|c|c|c|c|c|c|}
\hline
  s                      &  p       & l       & n         & t         & st         & q  & m               & mm       
\tabularnewline
\hline \hline
1                      &  1      & 1       & 0         & 0         & 0         & 1  & 1               & 0       
\tabularnewline
\hline 
1                      &  1      & 1       & 0         & 0         & 1         & 1  & 1               & 0       
\tabularnewline
\hline 
..                      &  ..      & ..       & ..         & ..         & ..         & ..  & ..               & ..       
\tabularnewline
\hline
\end{tabular} 
\label{tab:tableScompleteenumeration} 
\end{table}

\emph{(b) ``One Table per Feature'' Representation}. With an increasing configuration space size, an implicit representation is needed since an enumeration of all possible feature combinations is impossible or would lead to serious performance issues. An alternative approach to represent the configuration space of a feature model is depicted in Table \ref{tab:tableSimplicit} where each feature of the feature model in Figure \ref{fig:FMSurveySoftware} is represented by a separate table, for example, feature $p$ (payment) is represented by the table $p$ including the attribute $val$ and corresponding table entries $0$ and $1$ expressing feature exclusion or inclusion. In this context, a conjunctive query returning a feature model configuration could be formulated as follows: $F_{[s.val=1,p.val=1,n.val=1,mm.val=0]}\{s.val .. mm.val\}$ where $F$ is regarded as implicit representation of the Cartesian product $s \times p \times l \times n \times t \times st \times q \times m \times mm$. On the level of \textsc{SQL}, a corresponding query is: \textsc{SELECT} \emph{s.val.. mm.val} \textsc{FROM} \emph{s .. mm} \textsc{WHERE} \emph{s.val=1 and p.val=1 and n.val=1 and mm.val=0}. 

\emph{Example Query Optimizations}. Possibilities of improving the performance of such queries are (a) to reduce the domain of the root feature to \emph{true} (1), (b) to reduce the domain of dead features to \emph{false}, and (c) to reduce the domain of false optional features to \emph{true} -- for related details on feature model analysis, we refer to Benavides et al. \cite{BeSeRu2010}.

\begin{table}[ht]
\centering 
\caption{Implicit configuration space description where each feature is represented by a single table with one attribute $val$ and the table entries $0$ and $1$.} 
\begin{tabular}{|c|c|c|c|c|c|c|c|c|c|c|}
\hline
  table & s                      &  p       & l       & n         & t         & st         & q  & m               & mm       
\tabularnewline
\hline 
 attribute & val                      &  val       & val       & val         & val         & val         & val  & val               & val       
\tabularnewline
\hline \hline
domain & {0,1}                         &  {0,1}        & {0,1}        & {0,1}          & {0,1}          & {0,1}           & {0,1}   & {0,1}               & {0,1}          
\tabularnewline
\hline
\end{tabular} 
\label{tab:tableSimplicit} 
\end{table}



\emph{(c)``Local Consistency'' based Representation}. An alternative to the ``One Table per Feature'' representation is to use tables for expressing local consistency properties. For example, the constraint \emph{AB testing} (\emph{t}) requires \emph{statistics} (\emph{st}) can be represented as corresponding consistency table \emph{t-st} (see Table \ref{tab:tableSimplicitlocal}) expressing all possible combinations of \emph{t} and \emph{st} taking into account solely the \emph{requires} relation between the two features.

\begin{table}[ht]
\centering   
\caption{Implicit configuration space  representing locally consistent feature assignments ($a_i$), e.g., in table \emph{t-st}.} 
\begin{tabular}{|c|c|c|c|c|c|c|c|c|c|c|}
\hline
  feature & t                      &  st       
\tabularnewline
\hline 
$a_1$ & 1 & 1    
\tabularnewline
\hline 
$a_2$ & 0 & 1      
\tabularnewline
\hline
$a_3$ & 0 & 0      
\tabularnewline
\hline
\end{tabular} 
\label{tab:tableSimplicitlocal} 
\end{table}

\begin{table}[ht]
\centering   
\caption{Locally consistent feature assignments $a_i$ of \emph{excludes} relationship between feature $n$ and $t$ represented by table \emph{n-t}.} 
\begin{tabular}{|c|c|c|c|c|c|c|c|c|c|c|}
\hline
  feature & n                      &  t       
\tabularnewline
\hline 
$a_1$ & 0 & 1    
\tabularnewline
\hline 
$a_2$ & 1 & 0      
\tabularnewline
\hline
$a_3$ & 0 & 0      
\tabularnewline
\hline
\end{tabular} 
\label{tab:tableSimplicitlocalanothertable} 
\end{table}

Since Tables \ref{tab:tableSimplicitlocal} and  \ref{tab:tableSimplicitlocalanothertable} include the feature $t$, we need to assure that in the final configuration, $t$ has exactly one value -- this can  be achieved be integrating  join conditions into the conjunctive query. In \textsc{SQL}, a corresponding query is: \textsc{SELECT} \emph{n-t.n,n-t.t,t-st.st} \textsc{FROM} \emph{n-t,t-st} \textsc{WHERE} \emph{n-t.t=t-st.t}. Using this representation, feature names are regarded as table attributes, for example, feature $t$ (\emph{AB testing}) is regarded as an attribute of  table \emph{n-t} as well as \emph{t-st}.

\emph{Example Query Optimizations}. Following the concept of \emph{k-consistency} in constraint solving \cite{handbookconstraintprogamming2006}, the amount of tuples in a constraint table (e.g., \emph{n-t}) can be reduced by taking into account further feature model constraints. In the extreme case, this would lead to global consistency \cite{handbookconstraintprogamming2006} which assures that each constraint table contains only tuples (representing feature value combinations) that can be part of at least one configuration. For complex feature models (FM), global consistency is impossible since this would basically require the enumeration of all possible configurations.

Summarizing, there are three basic options for representing FM configuration tasks as conjunctive queries: (a) enumerating all possible configurations (see Table \ref{tab:tableScompleteenumeration}), (b) implicit representation with one table per feature (see Table \ref{tab:tableSimplicit}), or (c) using tables to represent locally consistent features settings.

\section{New Approaches to Conflict Resolution}\label{sec:advancedconflictresolution}

In configuration scenarios, the conjunctive query $F_{[c]}S$ could result in the empty set. Such situations can occur, for example, if the feature model is void, i.e., no solution can be identified or the given set of customer requirements induces an inconsistency with the feature model constraints \cite{Felfernigetal2018,Felfernig2012}. In such situations, we need to identify minimal sets of constraints (in the feature model or the customer requirements) that have to be adapted in order to restore consistency. In the context of feature model configuration, there exist a couple of approaches supporting corresponding feature model diagnosis operations \cite{Benavidesetal2013,Hentze2021Vamos}. An alternative to the application of diagnosis algorithms is to directly exploit information included in table-based representations. An example thereof is depicted in Table \ref{tab:tablediagnoses}. Assuming a set of customer requirements $cr$, we are able to directly compare these requirements with a set of potential configurations. If $cr$ is \emph{inconsistent} with a configuration $conf_i$, those feature inclusions (exclusions) of $cr$ inconsistent with $conf_i$ represent a diagnosis $\Delta$ where $cr - \Delta$ is consistent with $conf_i$. 

\begin{table}[h!]
\centering 
\caption{Direct derivation of diagnoses.} 
\begin{tabular}{|c|c|c|c|c|c|c|c|c|c|c|}
\hline
  feature & s                      &  p       & l       & n         & t         & st         & q  & m               & mm       
\tabularnewline
\hline \hline
$cr$ & 1                        &  1       & 0       & 1         & 1         & 1          & 1  & 1               & 0  \tabularnewline
\hline   
$conf_1$ & 1                        &  1       & 0       & 1         & \bf0         & 1          & 1  & 1               & 0    
\tabularnewline
\hline
$conf_2$ & 1                        &  1       & \bf1       & \bf0         & 1         & 1          & 1  & 1               & 0 
\tabularnewline
\hline
..                      &  ..      & ..       & ..         & ..         & ..         & ..  & ..               & ..       & ..    
\tabularnewline
\hline
\end{tabular} 
\label{tab:tablediagnoses} 
\end{table}

This way, we are able to determine diagnoses without the need of applying corresponding diagnosis algorithms. In the example depicted in Table \ref{tab:tablediagnoses}, we are able to identify two potential diagnoses: $\Delta_1=\{t=1\}$ and $\Delta_2=\{l=0,n=1\}$. Applying $\Delta_1$ basically means to inform the user about a potential exclusion of feature $t$ which could then lead to a consistent configuration.


\section{Performance Analysis}\label{sec:performanceanalysis}


We compared the performance of three approaches for representing FM configuration tasks as conjunctive query with constraint solving on the basis of four real-world feature models selected from the \textsc{S.P.L.O.T.} feature model repository \cite{mendonca2009}. Table \ref{tab:table1} provides an overview of selected feature models. Due to space complexity, not all configurations could be determined for \emph{WebArch} within reasonable time limits.

For each feature model, we randomly synthesized\footnote{To ensure the reproducibility of the results, we used the seed value of 141982L for the random number generator.} and collected 25,000 user requirements that cover 40-60\% of the leaf features in the feature model. We applied the systematic sampling technique~\cite{mostafa2018} to select 10 \textit{no-solution} user requirements and 10 user requirements with at least one solution. In Table \ref{tab:table4}, each setting shows the average runtime of the corresponding approach after executing the queries on the basis of these 20 user requirements. We used \textsc{Choco Solver}\footnote{choco-solver.org} as a reasoning solver and \textsc{HSQLDB}\footnote{hsqldb.org} as an in-memory relational database management system. All reported experiments were run with an Apple M1 Pro (8 cores) with 16-GB RAM, and an HSQLDB maximum cache size of 4GB. 

Table \ref{tab:table4} shows the results of this evaluation of selected feature models represented as (a) an explicit enumeration of \emph{all configurations}, (b) an implicit representation of the feature model configuration space (\emph{one table per feature}), (c) an implicit representation where individual tables represent locally consistent feature assignments (\emph{one table per constraint}), and (d) constraint satisfaction problem (CSP). This initial evaluation shows similar runtimes for small feature models and significantly longer runtimes for conjunctive query based approaches in the case of more complex models. Basically, these initial results of our performance evaluation show the applicability of conjunctive query based approaches to the task of feature model configuration.

\begin{table}[h]
\caption{Properties of selected  feature models (IDE=IDE product line, DVS=digital video system, DELL=DELL Laptop feature model, WebArch=web architectures).} 
\label{tab:table1} 
\centering 
\begin{tabular}{|l|c|c|c|c|c|c|c|c|c|c|c|c|c|}
\hline
feature model & IDE & DVS & DELL & WebArch \tabularnewline \hline \hline
\#features & 14 & 26 & 47 & 77  \tabularnewline  \hline
\#leaf features & 9 & 16 & 38 & 46  \tabularnewline  \hline
\#hierarchical constraints & 11 & 25 & 16 & 65  \tabularnewline  \hline
\#cross-tree constraints & 2 & 3 & 105 & 0 \tabularnewline  \hline
\#configurations & 80 & 22,680 & 2,319 & - \tabularnewline  \hline
\end{tabular} 
\end{table}

\begin{table}[h]
\caption{The average runtime (\textit{msec}) of conjunctive query and constraint-based feature model configuration. }  
\label{tab:table4} 
\centering 
\begin{tabular}{|l|c|c|c|c|c|c|}
\hline
feature model & IDE & DVS & DELL & WebArch \tabularnewline \hline \hline
all configurations & 0.78 & 9.43 & 6.49 & - \tabularnewline  \hline
one table per feature  & 0.49 & 0.45 & 2.53 & 327.75 \tabularnewline  \hline
one table per constraint  & 0.51 & 0.78 & 3.32 & 294.55 \tabularnewline  \hline
CSP & 0.73 & 0.78 & 1.09 & 1.1 \tabularnewline  \hline
\end{tabular} 
\end{table}

\section{Threats to Validity \& Open Issues}\label{sec:validitythreats}
In this paper, we have shown how to apply conjunctive queries to the identification of feature model configurations. We have provided results of an initial performance analysis of explicit and implicit table-based representations of feature models. We have compared the performance of conjunctive (database) queries with corresponding constraint solver based implementations. Initial results are promising and show the applicability of our approach in terms of search efficiency. We are aware of the fact that further in-depth evaluations with industrial datasets and different types of knowledge representations are needed to gain further insights into the applicability of our approach. We also sketched alternative approaches to deal with inconsistencies in feature model configuration scenarios. We are aware that related evaluations are needed to be able to estimate in detail the potential improvements in terms of runtime performance of diagnosis and conflict search.

\section{Conclusions and Future Work}\label{sec:conclusionsandresearchissues}

We have introduced a conjunctive query based approach to feature model (FM) configuration. In this context, we have compared the performance of three alternative conjunctive query based knowledge representations. With this, we provide an alternative to SAT solving, constraint solving, and ASP which can reduce overheads due to new technology integration. Initial results of our performance analysis are promising, however, we are aware of the fact that  technologies such as SAT solving can outperform conjunctive queries. In addition, we have sketched how to exploit table-based representations for efficient conflict resolution. 

There are a couple of open issues for future work. First, we will extend our performance analyses to comparisons with a broader range of industrial feature models. We will also include SAT solving and ASP in our evaluations. We also plan to analyze the potentials of combining different worlds, for example, we will analyze to which extent approaches from constraint solving (e.g., forward checking) can help to improve the efficiency of conjunctive queries. Vice-versa, we will analyze to which extent search techniques from relational databases (and also machine learning) can help to further advance the field of SAT solving, constraint solving, and ASP. Finally, we want to analyze the properties of phase transitions, i.e., when to move from an explicit to an implicit search space representation.\\

\noindent
\textbf{Acknowledgments.} The work presented in this paper has been developed within the research project \textsc{ParXCel} (\emph{Machine Learning and Parallelization for Scalable Constraint Solving}) which is funded by the Austrian Research Promotion Agency (FFG) under the project number $880657$.

%
%
\bibliographystyle{splncs04}
\bibliography{bibliography}

\end{document}